\def\year{2019}\relax
\documentclass[letterpaper]{article} 
\usepackage{aaai19}  
\usepackage{times}  
\usepackage{helvet}  
\usepackage{courier}  
\usepackage{url}  
\usepackage{graphicx}  
\usepackage{subcaption}
\usepackage{graphicx}
\usepackage{amsfonts}
\usepackage{amsmath}
\usepackage{breqn}

\newcommand{\citet}[1]{\citeauthor{#1}~\shortcite{#1}}
\newcommand{\citep}{\cite}

\begin{document}
%
\title{Hybrid Reinforcement Learning with Expert State Sequences}
\author{Xiaoxiao Guo, Shiyu Chang, Mo Yu, Gerald Tesauro, Murray Campbell \\
IBM Research AI \\
\{xiaoxiao.guo, shiyu.chang\}@ibm.com, \{yum, gtesauro, mcam\}@us.ibm.com}
\maketitle

\begin{abstract}
Existing imitation learning approaches often require that the complete demonstration data, including sequences of actions and states, are available.  In this paper, we consider a more realistic and difficult scenario where a reinforcement learning agent only has access to the state sequences of an expert, while the expert actions are unobserved.  We propose a novel tensor-based model to infer the unobserved actions of the expert state sequences.  The policy of the agent is then optimized via a hybrid objective combining reinforcement learning and imitation learning.  We evaluated our hybrid approach on an illustrative domain and Atari games.  The empirical results show that (1) the agents are able to leverage state expert sequences to learn faster than pure reinforcement learning baselines, (2) our tensor-based action inference model is advantageous compared to standard deep neural networks in inferring expert actions, and (3) the hybrid policy optimization objective is robust against noise in expert state sequences.  
\end{abstract}

\section{Introduction}

Human expert behavioral data are widely used for policy learning in sequential decision-making tasks~\cite{schaal1999imitation,argall2009survey}.  One of the most effective
paradigms is {\em imitation learning}, where a policy is trained via direct supervision to clone expert behaviors~\cite{pomerleau1989alvinn,ross2011reduction}.  Imitation learning generally requires both observable states and actions as input.  However, expert actions are
often unavailable or not directly usable.  Literature has considered scenarios where the expert and the imitation learner may have different viewpoints~\cite{liu2017imitation}, temporal resolution or action sets~\cite{yu2018one}.  In such cases, cloning behavior directly is not an option.    
How to leverage such data to facilitate learning is a realistic and challenging problem.

\begin{figure}
\centering
\includegraphics[width=0.95\linewidth]{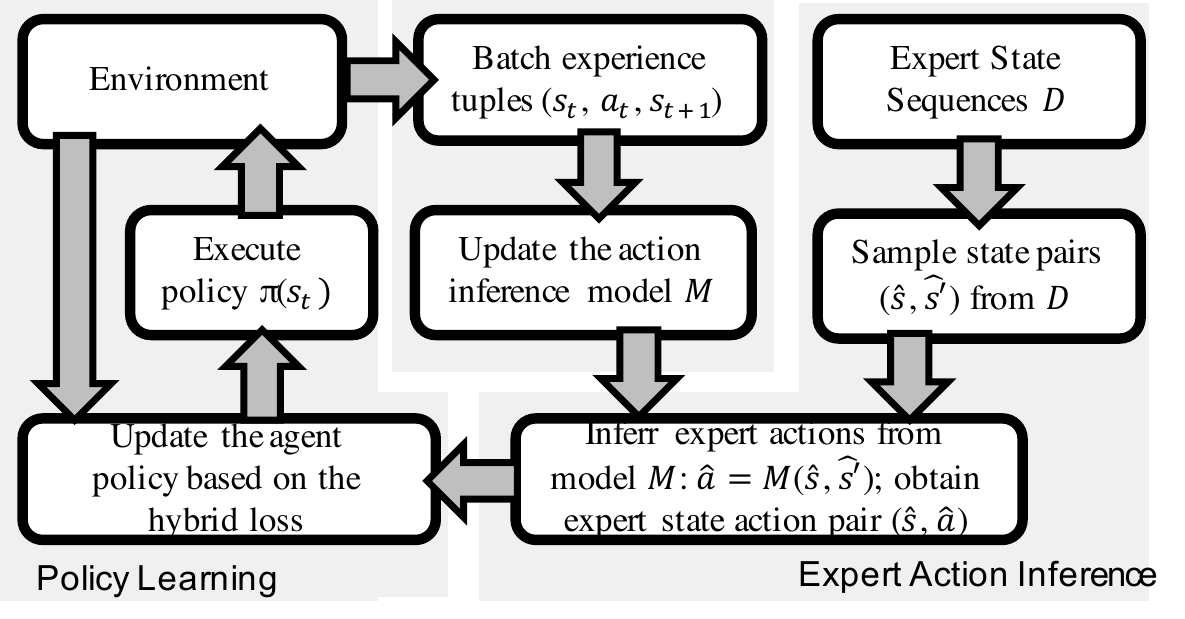}
\caption{The proposed hybrid reinforcement learning with expert state sequences framework. \label{fig:procedure}}
\end{figure}

In this paper, we investigate a novel learning scenario where an agent learns from both its own experience and state-trajectory-only expert demonstrations.  We propose an iterative learning framework as follows, illustrated in Figure~\ref{fig:procedure}.   We first learn a novel tensor-based action inference model as the learning agent interacts with the environment.  Our model enforces a 
duality for 
consistent learning as the inferred action from two consecutive states reconstructs the latter state.  Then, upon observation of demonstration without actions, the learned dynamics is used to infer the missing actions.  Finally, we improve the learning policy by jointly considering the imitation performance and rewards from environment interaction via Advantage Actor-Critic (A2C)~\cite{baselines}.  

To demonstrate the effectiveness of the proposed iterative learning process, we conduct experiments on the Taxi domain as well as eight commonly used Atari games.  The experimental results confirm  a faster convergence rate of the proposed framework compared to advantage actor-critic alone, and a better policy compared to behavioral cloning from observations (BCO)~\cite{torabi2018behavior}, which only considers a similar action inference approach for behavioral cloning but ignores potential reward signals.  We additionally show that our framework is robust against noisy expert state trajectories, and works well even when the number of demonstrations is limited. 

\section{Related Work}
\paragraph{Imitation Learning} The tasks of learning from demonstrations and imitation learning
have attracted considerable research attention from many fields in machine learning.
The approaches are roughly divided into two groups,
behavioral cloning and Inverse Reinforcement Learning (IRL). Survey articles include~\cite{schaal1999imitation,billard2008robot,argall2009survey}.  Behavioral Cloning (BC) uses supervised learning, where the learner directly regresses onto the policy of the expert~\cite{pomerleau1989alvinn,ross2011reduction,liu2017imitation}.
This requires observation-action tuples, and cannot be applied when actions are absent.  On the other hand, IRL methods aim to infer the goal of an expert,
expressed as a reward function ~\cite{ng2000algorithms,abbeel2004apprenticeship,ziebart2008maximum,levine2011nonlinear,borsa2017observational}. 
This can then be used for RL to recover a policy~\cite{ratliff2006maximum,ramachandran2007bayesian}.  Contemporaneous work~\cite{stone2018ijcai} extends BC to Behavioral Cloning from Observations (BCO), a setting where expert state-transitions are observed but actions are not observed. This differs from our setting where we hybridize BCO with RL. 

\paragraph{Hybrid RL with Expert Actions} Several work has
focused on combining standard RL with supervised training on expert actions, for example, by
alternating steps of RL and IRL to obtain more accurate estimates~\cite{ho2016generative,finn2016guided}.  Although both lines of research were successfully applied to a variety of tasks, they almost always assume that the state-action trajectories of experts are in the same space as the learner observed.  As argued in some recent work~\cite{stadie2017third,duan2017one,liu2017imitation}, such an assumption is restrictive and unrealistic.  Instead, they proposed to discover the transformations between the learner and the teacher state space.
Additional recent work combines human expert data and RL in policy learning. \citet{gilbert2015reducing}, \citet{lipton2016efficient} and~\citet{xxyyzz} store expert state-action pairs in a replay buffer to facilitate learning.  \citet{hosu2016playing} utilize expert state-action pairs to facilitate action value learning.  \citet{subramanian2016exploration} leverage human data for efficient exploration. \citet{hester2017learning} combine several approaches and demonstrates superior performance on Atari games. ~\citet{nair2017rldemo} reported significant progress in training robotic tasks using a combination of RL and a small set of human demonstrations. 
Our method differs from all these approaches in that we do not assume expert actions are available.

Very recent work starts to investigate leveraging expert state sequences only to accelerate imitation learning.
\citet{aytar2018playing} utilize state only demonstration data to address the hard exploration issue for Atari games. \citet{zhu2018reinforcement} leverage a small amount of demonstration data to assist a reinforcement learning agent for robotic manipulation tasks.

\paragraph{Model-based RL} Researchers have known for decades that learning a domain model concurrently with learning a behavior policy can significantly improve over model-free RL~\cite{sutton91dyna}. ~\citet{Chebotar-2017-964} recently demonstrated much better sample efficiency using model-based RL for robotics tasks, without requiring demonstrations.  ~\citet{oh2015action} and~\citet{machado2018eigenoption} trained deep neural network models to predict the next frame or successor representation in Atari games with good effect. They applied element-wise multiplications on the state and action embeddings to obtain the embedding of the next state, which could be viewed as a special case of our model.

\section{Sequential Decision Making with Expert State Sequences}
We formulate the sequential decision-making task as a Markov decision process (MDP).  An MDP is a tuple $<\mathcal{S}, \mathcal{A}, P, R, \gamma>$ where $\mathcal{S}$ is the state space and $\mathcal{A}$ is the action space. $P: \mathcal{S} \times \mathcal{A} \times \mathcal{S} \rightarrow [0,1]$ is the state transition function and $P(s' | s,a)=\textrm{Pr}(s'|s,a)$ is the probability that the next state is $s'$ given a current state $s$ and action $a$ is taken. $R: \mathcal{S} \times \mathcal{A} \rightarrow \mathbb{R}$ is the reward function with $R(s,a)$ being the expectation of immediate rewards, $r(s,a)$, of taking action $a$ in state $s$.  $\gamma$ is a temporal discounting factor.  A stochastic policy $\pi: \mathcal{S} \times \mathcal{A} \rightarrow [0,1]$ specifies the action to take in states. A state value function is defined as the expected sum of discounted rewards following a policy from a state: $V^{\pi}(s)=\mathbb{E}[\sum^{\infty}_{t=1}\gamma^{t-1} r_{t}| s_{0}=s,\pi]$. Similarly, a state-action value function is defined as $Q^{\pi}(s,a)=\mathbb{E}[\sum^{\infty}_{t=1}\gamma^{t-1} r_{t}| s_{0}=s, a_{0}=a, \pi]$. The optimal policy $\pi^{*}$ has action value function $Q^{*}=\max_{\pi}Q^{\pi}$. Taking actions greedily with respect to $Q^{*}$ yields the optimal policy $\pi^{*}$.  

In addition, we assume a set of state sequences (with unknown actions) demonstrated by an expert is given.  We denote expert state sequences as a set of $N$ state pairs,  $\mathcal{D} = \{ (\hat{s}_{i}, \hat{s}'_{i}) \}_{i=1}^{N}$, where $\hat{s}'_{i}$ is the next state of $\hat{s}_{i}$. 
Note that we do not assume that states are consecutive across different pairs, \emph{i.e.}, $\hat{s}'_{i}$  is not necessarily equivalent to $\hat{s}_{i+1}$.  We aim to design a flexible framework that can accommodate dataset $\mathcal{D}$ in various formats.

\section{Hybrid Reinforcement Learning with Expert State Sequences}
To utilize the expert state sequences, our method learns a model of the environment to infer the missing expert actions from consecutive expert states. The inferred actions combined with the expert states are then utilized to provide additional supervision on the policy of our agent via behavioral cloning. The learning paradigm of our agent is illustrated in Figure~\ref{fig:procedure}. The agent interacts with the environment following its current policy $\pi(.;\theta)$ as traditional RL agents. $\theta$ denotes the learning parameters of the policy $\pi$. When the agent interacts with the environment, the state-action-next-state tuples $\{(s,a,s')\}$ of its experience are collected to additionally train an action inference model $\mathcal{M}: \mathcal{S} \times \mathcal{S} \rightarrow \mathcal{A}$, which maps the state-next-state pair into an action. The details of the model $\mathcal{M}$ will be provided in the Low-Rank Tensor Formulation section. We sample a batch of consecutive expert state pairs $\{ (\hat{s}_{i},\hat{s}_{i}')\}$ and apply the action inference model $\mathcal{M}$ to obtain the action estimate $\hat{a}_{i}=\mathcal{M}(\hat{s}_{i},\hat{s}_{i}')$ for each expert state pair $ (\hat{s}_{i},\hat{s}_{i}')$. The action estimate $\hat{a}_{i}$ and the expert state $\hat{s}_{i}$ are combined as a batch of expert state-action pairs $\{ (\hat{s}_{i}, \hat{a}_{i}) \}$ to optimize the policy of the agent via behavioral cloning. The agent also applies RL to optimize its policy simultaneously.

In the rest of this section, we provide details of the action inference model $\mathcal{M}$, followed by the hybrid training objective combining behavioral cloning and RL.  

\subsection{Modeling State Transition Dynamics and Action Inference}

Traditional model-based RL usually estimates the forward dynamics of the stochastic state transitions of the environment as $P^{f}: \mathcal{S} \times \mathcal{A} \times \mathcal{S} \rightarrow [0,1]$ with $P^{f}(s'|s,a)$ being the likelihood of the next states $s'$ conditioned on the current state $s$ and action $a$. Ideally the action inference model should be consistent with such model-based RL approach. Specifically, the output of the action inference model should be the action maximizing the likelihood of the observed state transition. 
Such approach may require $|\mathcal{A}|$-time computation of $P^{f}(.)$ in order to find the best action output. A more computation-friendly solution would be to obtain a consistent view of the state transitions which directly estimates the likelihood of actions conditioned on two consecutive states $P^{i}: \mathcal{S} \times \mathcal{S} \times \mathcal{A} \rightarrow [0,1]$, with $P^{i}(a | s,s')$ being the likelihood of action $a$ for the state pair $(s,s')$.  $\mathcal{M}(s,s')=\arg\max_{a} P^{i}(a|s,s')$. Tensors offer natural representation of the joint model of $P^{f}$ and $P^{i}$. A tensor is able to compute the two views of the state transitions effectively. One technical contribution of this paper is a tensor-based action inference model to maintain the two views of the environment dynamics consistently and simultaneously.      

\subsection{Low-Rank Tensor Formulation}
\paragraph{Motivation in Small Domains}
For an MDP with state space $\mathcal{S}$ and action space $\mathcal{A}$, $P^{f}$ and $P^{i}$ can be represented as tensor multiplication on a shared tensor $\mathcal{T} \in \mathbb{R}^{|\mathcal{S}| \times |\mathcal{A}| \times |\mathcal{S}|}$ where $\mathcal{T}(s,a,s')$ stores the count number, $c(s,a,s')$,  of the tuple $(s,a,s')$ in the agent's experience. Then the maximum likelihood estimate of $P^{f}(s'|s,a)=c(s,a,s')/ \sum_{s'}c(s,a,s')$ and $P^{i}(a|s,s')=c(s,a,s')/ \sum_{a}c(s,a,s')$ could be represented as:

\begin{equation*}
\begin{split}
P^{f}(s' | s,a) &= h_{sa}[s'] / ||h_{sa}||_{1} \\
h_{sa} = c(s,a,:) &= \mathcal{T} \times_{1} \mathbf{1}(s) \times_{2} \mathbf{1}(a)
\end{split}
\end{equation*}

\begin{equation*}
\begin{split}
P^{i}(a | s,s') &= h_{ss'}[a] / ||h_{ss'}||_{1} \\
h_{ss'} = c(s, :, s') &= \mathcal{T} \times_{1} \mathbf{1}(s) \times_{3} \mathbf{1}(s')
\end{split}
\end{equation*}
where $\mathbf{1}(i)$ is a one-hot vector at location $i$, $||.||_{1}$ is the $L_{1}$ norm of a vector, and $\times_{m}$ denotes the \textbf{mode-m product}, defined as $ (\mathcal{T}\times_{m} W)_{i_{1},...,j,...,i_{K}} = \sum_{i_{m}=1}^{d_m} \mathcal{T}_{i_{1}, ..., i_{m}, ..., i_{K}}W_{i_{m}, j}$. Note that $c(s,a,s')=\mathcal{T}\times_{1} \mathbf{1}(s)\times_{2}\mathbf{1}(a)\times_{3}\mathbf{1}(s')$, and $c(s,a,s')$ can be viewed as a score of the tuple ($s,a,s'$) from tensor multiplications.
$h_{sa}$ is a vector of length $|\mathcal{S}|$ with $h_{sa}[s']=c(s,a,s')$ and $h_{ss'}$ is a vector of length $|\mathcal{A}|$ with $h_{ss'}[a]=c(s,a,s')$. $P^{f}$ and $P^{i}$ are normalized vectors of $h_{sa}$ and $h_{ss'}$. 
By organizing the count numbers of the tuples ($s,a,s'$) into tensor, $P^{f}$ and $P^{i}$ could be modeled jointly by the same tensor $\mathcal{T}$ in the tabular cases. But for MDPs with large state space, such tensor would be both memory-demanding and computationally expensive.    

\begin{figure}
\centering
\includegraphics[width=0.95\linewidth]{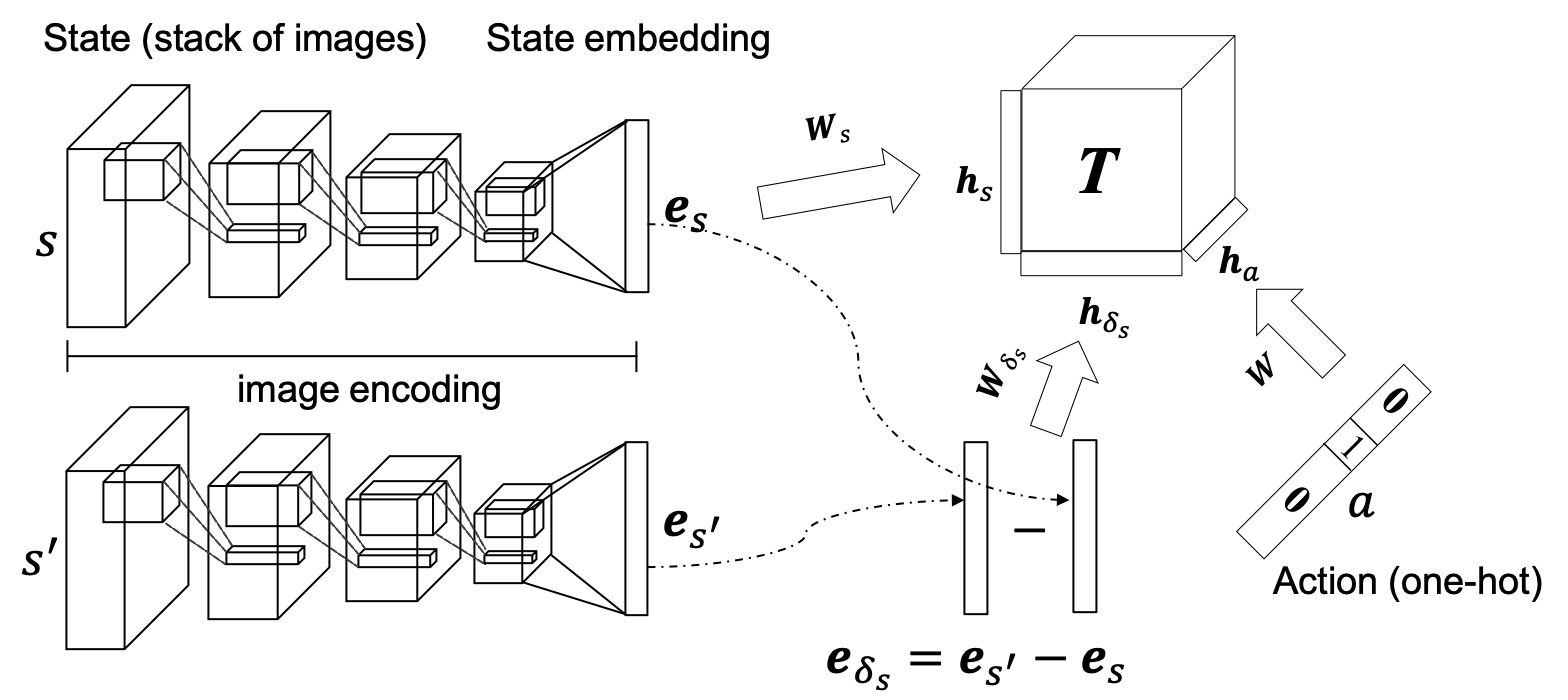}

\caption{Tensor formulation of state-transition modeling for playing Atari games. A state consists of a window of frames, which are images from the game screen. Those images pass a convolutional neural network and result in a state embedding (left part). Given a tuple $(s, a, s')$, our model predicts a score by multiplying a tensor $\mathcal{T}$ with the hidden representations of $s, a, \delta_s$, i.e. $\mathbf{h}_s, \mathbf{h}_a, \mathbf{h}_{\delta_s}$ (right part). The matrices $\mathbf{W_{s}}, \mathbf{W}_{\delta s}$ and $\mathbf{W}$ map state embeddings $\mathbf{e}_s$ and $\mathbf{e}_{s'} - \mathbf{e}_{s}$ and the one-hot encoding of action $a$ to vectors.\label{fig:low_rank}}
\end{figure}

We reduce the tensor computation via the following approaches: 
(1) We reduce the dimensionality of the tensor by allowing the states embed in a lower dimension. We embed the state representation $\mathbf{e}_s \in \mathbb{R}^{s}$\footnote{For example, $\mathbf{e}_s $ could be a one-hot  vector from a lookup table for small state space domains, or an output vector from ConvNet to represent image inputs for video game domains.} to a lower dimension space $\mathbb{R}^{d_{s}}$ via a matrix $\mathbf{W_{s}} \in \mathbb{R}^{d_{s} \times s}$: $\mathbf{h}_{s}=\mathbf{W_{s}}\mathbf{e}_{s}$.  Instead of using the embedding of the next state $\mathbf{h}_{s'}$, we use the embedding for the difference between two states in tensor multiplications. The rationale in embedding the state differences instead of next states directly is that not all the information in the next states are relevant to the actions.  Moreover, the difference between two states could be embedded in a even lower dimensional space since not all state information are relevant to state transitions. The effects of actions are more related to the state differences. We denote the state difference embedding as $\mathbf{h}_{\delta s} = \mathbf{W}_{\delta s}(\mathbf{h}_{s'}-\mathbf{h}_{s}) \in \mathbb{R}^{d_{\delta s}}$ where $\mathbf{W}_{\delta s} \in \mathbb{R}^{d_{\delta s} \times d_{s}}$. Similarly, we embed the actions $\mathbf{h}_{a} \in \mathbb{R}^{d_a}$. Thus the score of the a tuple ($s,a,s'$) can be computed as $\textrm{score}(s,a,s')=\mathcal{T}\times_{1}\mathbf{h}_{s}\times_{2}\mathbf{h}_{a}\times_{3}\mathbf{h}_{\delta s}$ where $\mathcal{T}\in\mathbb{R}^{d_{s}\times d_a \times d_{\delta s}}$. 
Figure \ref{fig:low_rank} demonstrates the above formulation with an Atari game example.
Under this formulation, the predicted representation of the action and state difference is thus:
\begin{equation*}
\begin{split}
\hat{\mathbf{h}}_{\delta s} = \mathcal{T} \times_{1} \mathbf{h}_{s} \times_{2} \mathbf{h}_{a}, \\
\hat{\mathbf{h}}_{a} = \mathcal{T} \times_{1} \mathbf{h}_{s} \times_{3} \mathbf{h}_{\delta s} 
\end{split}
\end{equation*}

In order to have consistent score for the tuple ($s,a,s'$) using either the predicted embedding or the original embedding, part of the learning objective is to minimize the distance between the predicted representation and the original ones: $\min \mathbb{E}_{s,a,s'}\big[||\hat{\mathbf{h}}_{\delta s} -\mathbf{h}_{\delta s}||_{1} + ||\hat{\mathbf{h}}_{a} -\mathbf{h}_{ a}||_{1}\big]$.

(2) Even though $\mathcal{T}$ is now independent from the state space and the action space, it may still be computationally expensive. To further reduce the computation, first we introduce symmetry, $\mathcal{T}[:,k,:] = \mathcal{T}[:,:,k]$ with $d_{\delta s} = d_{a}=d$,  into the tensor by assuming the action and the difference in the state embedding between the current state and the next state could be embedded similarly when conditioned on the current state. Then we approximate the tensor slices $\mathcal{T}[:,k,:]$ as a sum of $R$ rank-1 matrices following ~\citet{li2017visual}:
\begin{equation*}
\mathcal{T}[:,k,:] = \sum_{r=1}^R \mathbf{M_r}[:,k] \otimes \mathbf{N_r}[:,k]^T. \label{low_rank_core_tensor}
\end{equation*}
where matrix $\mathbf{M_r} \in \mathbb{R}^{d_s \times d}$, $\mathbf{N_r} \in \mathbb{R}^{d \times d}$ and $\otimes$ denotes the outer product. $[:,k]$ denotes the $k$-th column of the matrix.
Thus each element $k$ of $\mathbf{\hat{h}}_a$ can be written as
$
\mathbf{\hat{h}}_a[k] = \sum_{r=1}^R (\mathbf{h}_s \mathbf{M_r}[:,k] ) \times (\mathbf{h}_{\delta s} \mathbf{N_r}[:,k])
$.
Letting $\odot$ denote Hadamard (point-wise) product, we have:
\begin{equation*}
\mathbf{\hat{h}}_a = \sum_{r=1}^R (\mathbf{h}_s \mathbf{M_r} ) \odot (\mathbf{h}_{\delta s} \mathbf{N_r}).\nonumber
\end{equation*}

With our symmetric $\mathcal{T}[:,:,k]=\mathcal{T}[:,k,:]$, the tensor slides $\mathcal{T}[:,:,k]$ could be approximated by the same $\mathbf{M_{r}}$ and $\mathbf{N_{r}}$, thus we have:
\begin{equation}
\mathbf{\hat{h}}_{\delta s} = \sum_{r=1}^R (\mathbf{h}_s \mathbf{M_r} ) \odot (\mathbf{h}_{a} \mathbf{N_r}).\nonumber
\end{equation}

This gives a computationally efficient dual state transition model $\mathcal{F}(s,x)=\sum_{r=1}^{R} (\mathbf{h}_{s} \mathbf{M_{r}}) \odot (\mathbf{h}_{x} \mathbf{N_{r}})$ to predict both $\hat{\mathbf{h}}_{\delta s}$ ($x$ is action) and    $\hat{\mathbf{h}}_{a}$ ($x$ is state-difference $\delta_s$). The predicted $\hat{\mathbf{h}}_{a}$ is then used to predict the action probability 
$P^{i}(.|s,s')=\textrm{SoftMax}(\mathbf{W_{a}}\mathbf{\hat{h}}_{a}+\mathbf{b}_a)$,
where $\mathbf{W}_{a} \in\mathbb{R}^{|\mathcal{A}|\times d}$ and $\mathbf{b}_{a} \in\mathbb{R}^{|\mathcal{A}|}$.

Similarly, the probability of the next state $s'$ is $P^{f}(.|s,a)=\textrm{SoftMax}(\mathbf{W}_{ s'}(\mathbf{\hat{h}}_{\delta s}+\mathbf{W'}_{\delta s}\mathbf{h}_{s})+\mathbf{b}_{s'})$, where $\mathbf{W}_{s'}\in\mathbb{R}^{|\mathcal{S}|\times d }$, $\mathbf{W'}_{\delta s}\in\mathbb{R}^{d \times d_{s}}$ and $\mathbf{b}_{s'}\in\mathbb{R}^{|\mathcal{S}|}$.
\footnote{Exactly following the tensor scoring function gives the probability estimation $P^{i}(.|s,s')=\textrm{SoftMax}(\mathbf{H}_{a}\mathbf{\hat{h}}_{a})$, where $\mathbf{H}_{a} \in\mathbb{R}^{|\mathcal{A}|\times d}$ is the concatenation of $\mathbf{h}_a$s for all $a$; and $P^{f}(.|s,a)=\textrm{SoftMax}(\mathbf{H}_{ s'}\mathbf{\hat{h}}_{\delta s})$, where $\mathbf{H}_{s'}\in\mathbb{R}^{|\mathcal{S}|\times d}$ is the concatenation of $(\mathbf{h}_{s'} + \mathbf{h}_{s})$s for all $s'$. However, as adopted by~\citet{li2017visual}, the proposed estimation
gives better empirical results. This is possibly because the classification objective benefits from more free parameters. Furthermore, $P^{f}$ can be replaced by reconstruction loss for large state space as used in~\citet{oh2015action}.}

\begin{figure}[!t]
\centering
\begin{subfigure}[b]{0.45\textwidth}
        \centering
        \includegraphics[width=\textwidth]{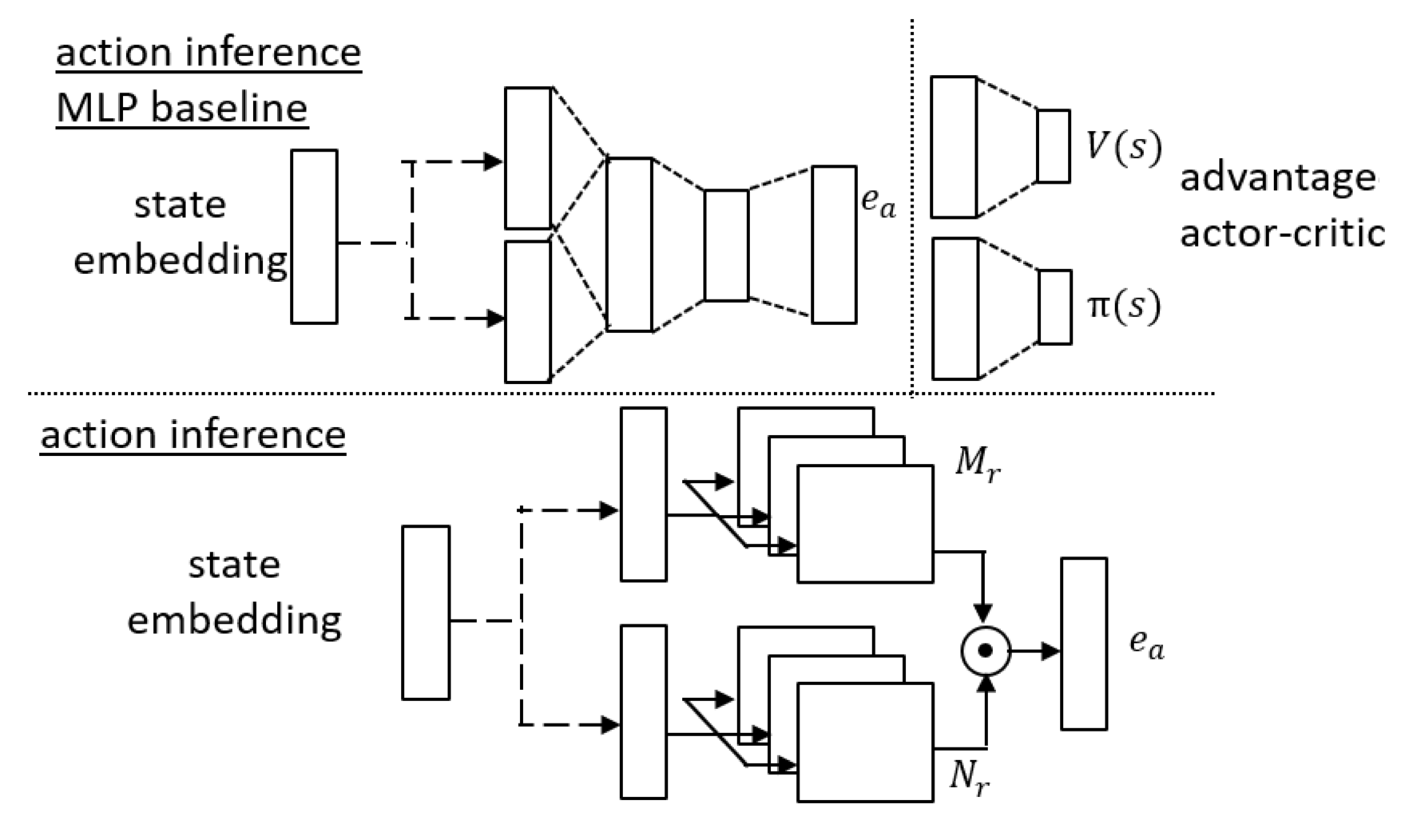}
\caption{\label{fig:net1}}
\end{subfigure}
\begin{subfigure}[b]{0.45\textwidth}
        \centering
        \includegraphics[width=\textwidth]{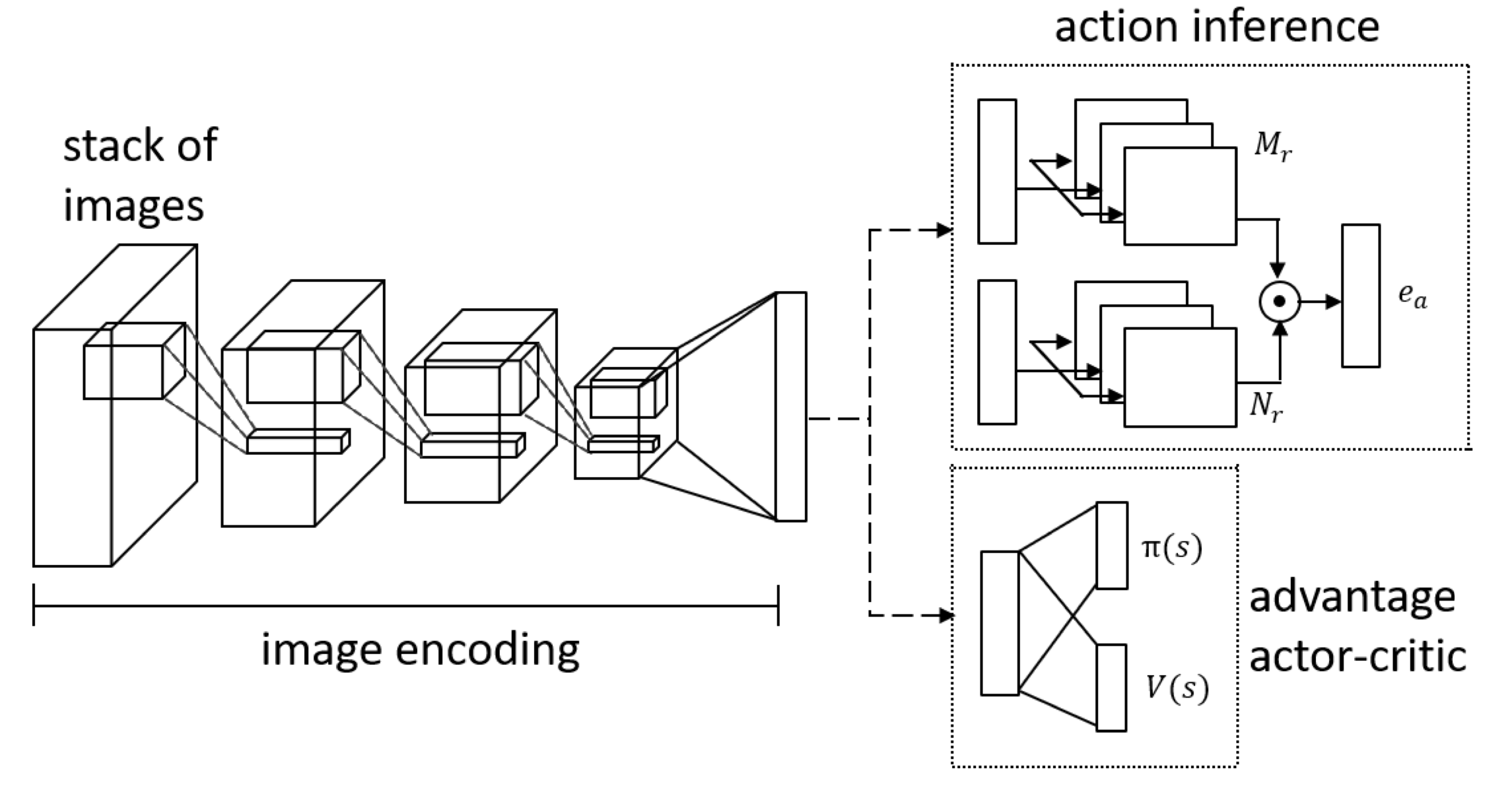}
\caption{\label{fig:net2}}
\end{subfigure}
\caption{Model architectures for (a)  Taxi and (b) Atari games. The Multi-Layer Perceptron (MLP) action inference baseline is also visualized in (a).}
\end{figure}

\paragraph{Learning Objective for the  State Transition Model}
The experience of the agent while interacting with the environment is used to optimize the dual state transition model $\mathcal{F}$. Since $\mathcal{F}$ is end-to-end trainable, we optimize $\mathcal{F}$ to maximize the likelihood of the tuple ($s,a,s'$) of the agent's own experience. 

The training objective is defined as follows:
\begin{dmath*}
\mathcal{L}^{\textrm{dual-model}} = \mathbb{E}_{(s,a,s')} \Big[ -\log P^{f}(s'|s,a) - \log P^{i} (a | s, s') 
+ \Vert\mathbf{h}_a - \mathbf{{\hat{h}}}_a\Vert_1 + \Vert\mathbf{h}_{\delta s} - \mathbf{{\hat{h}}}_{\delta s}\Vert_1 \Big] 
\end{dmath*}
In our learning scenario, only $P^{i}$ is relevant so we simplify the objective to model only the action inference part:  
\begin{dmath*}
\mathcal{L}^{\textrm{act}} = \mathbb{E}_{(s,a,s')} \Big[ - \log P^{i} (a | s, s') + \\
\Vert\mathbf{h}_a - \mathbf{{\hat{h}}}_a\Vert_1 + \Vert\mathbf{h}_{\delta s} - \mathbf{{\hat{h}}}_{\delta s}\Vert_1 \Big] 
\end{dmath*}
The learned $P^{i}$ is used to infer the actions given two consecutive expert states ($\hat{s}, \hat{s}'$): $\mathcal{M}(\hat{s},\hat{s}')=\hat{a}=\arg\max_{a}P^{i}(a|\hat{s},\hat{s}')$.

\begin{figure*}[!t]
\centering

\begin{subfigure}[b]{0.33\textwidth}~
        \centering
        \includegraphics[width=0.95\textwidth]{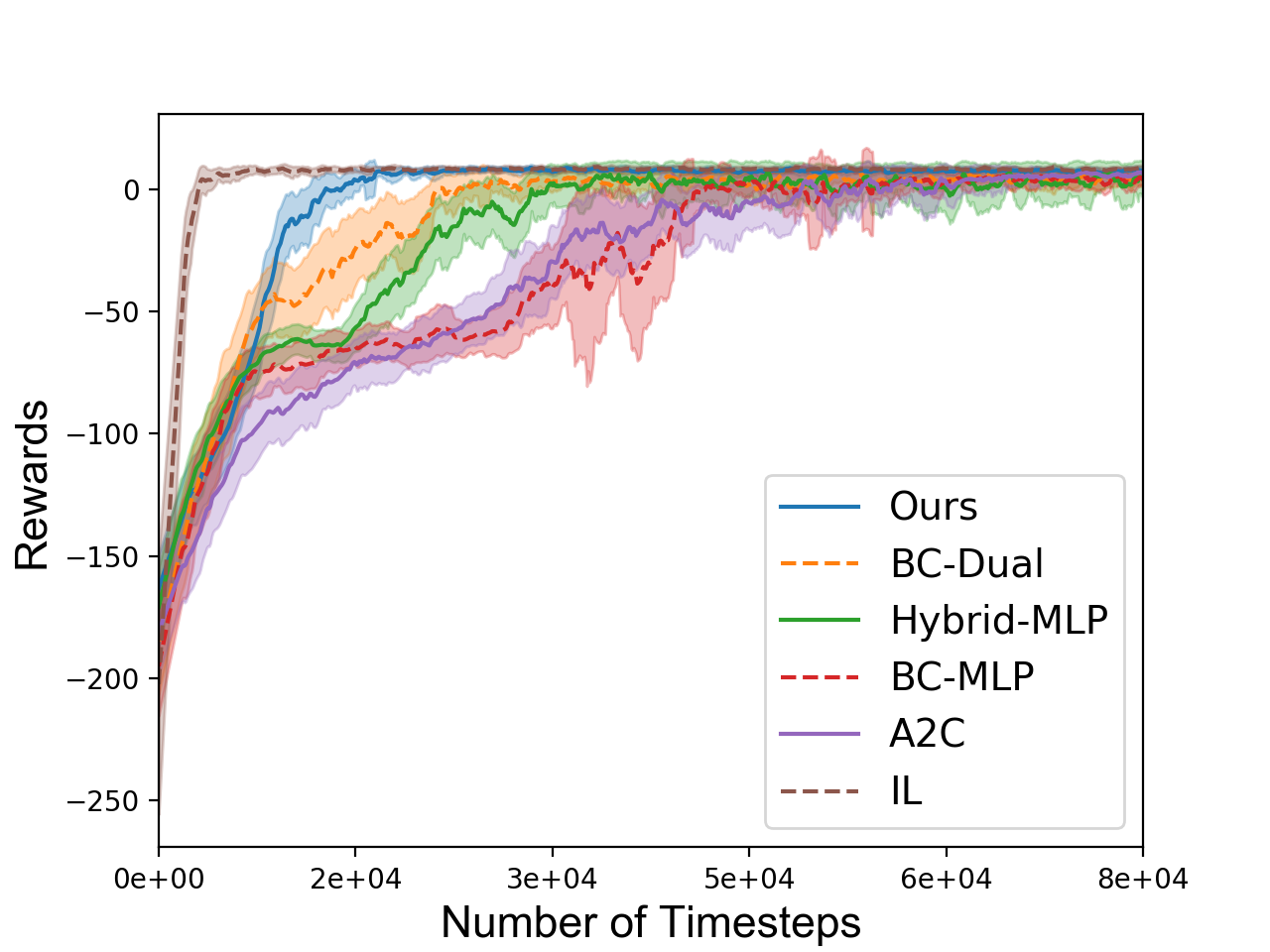}

\caption{ \label{fig:taxi-1}}
\end{subfigure}
\begin{subfigure}[b]{0.33\textwidth}~
        \centering
        \includegraphics[width=0.95\textwidth]{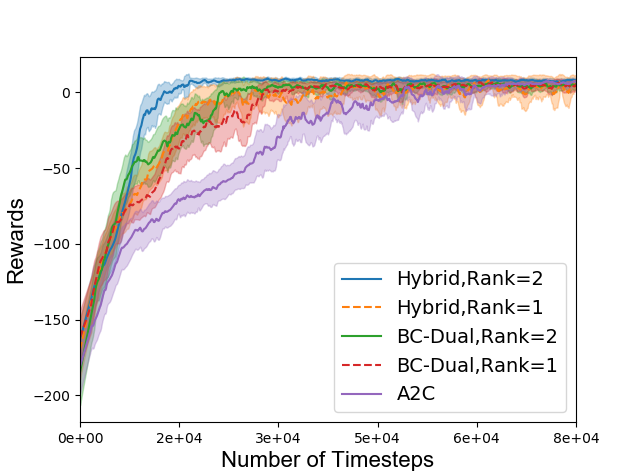}

\caption{\label{fig:taxi-2}}
\end{subfigure}
\begin{subfigure}[b]{0.33\textwidth}
        \centering
        \includegraphics[width=0.95\textwidth]{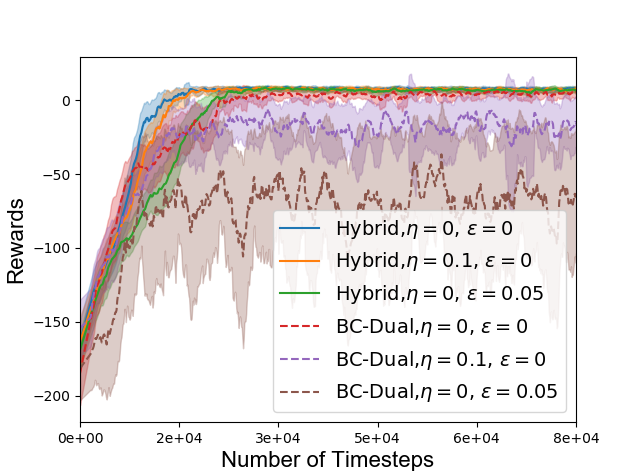}
\caption{\label{fig:taxi-3}}
\end{subfigure}
\caption{Learning curves of different agents on the Taxi domain. The figures show the average cumulative rewards across 80,000 time steps. The shadow regions represent the standard deviation of the average cumulative rewards. The figures are best viewed in colors. (a) The performance of our method and different baselines. BC represents agents only use the inferred actions on expert states to optimize the policies. MLP/Dual represent the architecture of the action inference model.   (b) The performance of our method and variant with different action inference model ranks. (c) The performance of our method varies with different types of noise from expert state sequences.  
\label{fig:taxi-all}}
\end{figure*}

\subsection{Hybrid Learning Objective}
The hybrid training objective of the policy $\pi$ combines the RL objective to maximize the expected sum of the discounted rewards and the imitation objective to maximize the likelihood of the inferred actions on the demonstrated states. Our RL method is Advantage Actor Critic (A2C). A2C learns the state value $V(s)$ by minimizing the squared advantage function values $A(s_{t})=\sum_{n=1}^{N} \gamma^{n-1} r_{t+n} + \gamma^{N} V(s_{t+N}) - V(s_{t})$. A2C optimizes the policy via policy gradient $ \mathbb{E} \big[ A(s) \nabla_{\theta} \log \pi(a|s) \big]$. Let $\theta$ denote the parameters of the policy $\pi$. The hybrid objective of our policy learning is:  
\begin{dmath}
\mathcal{U}^{\textrm{hybrid}}(\theta) = \mathbb{E}_{s,a}\Big[ A(s) \log \pi(a|s; \theta) +\alpha \mathcal{H}(\pi(.|s))\Big] + \mathbb{E}_{(\hat{s},\hat{s}') \sim \rho(\mathcal{D})}\Big[ \log \pi\big(\mathcal{M}(\hat{s},\hat{s}') | \hat{s};\theta\big) \Big] 
\end{dmath}
where $\rho()$ is a sampling distribution on the expert state pairs. It could be uniform or biased to match the state distribution of the agent's policy via curriculum learning. $\mathcal{H}(\pi(.|s))$ is the entropy of the policy for state s. 

\section{Empirical Evaluation}

To validate the proposed learning paradigm, we evaluate our proposed method on the Taxi domain~\cite{dietterich1998maxq} and eight Atari games from OpenAI Gym~\cite{openaigym}.  

\subsection{Taxi Domain}
We first evaluate our method on the Taxi domain. In addition, we analyze the performance of our proposed method when different types of noise exist in the expert state sequences and show that our hybrid learning approach is more robust compared to pure behavioral cloning from expert state sequences ~\cite{torabi2018behavior}. Last, we illustrate the parameter compression rate compared to the full tensor approach and analyze the parameter sensitivity. 

\subsubsection{Experiment Setup}  
As shown in Figure~\ref{fig:net1}, our agent architecture consists of two main components: A2C and the action inference model. A2C uses two forward step estimation for the advantage function. The state is represented as a one-hot vector of length 500 for both the actor and critic. The policy and state values are computed via separated linear transformations. No parameters are shared between the actor and the critic. The action inference model first projects current states and next states to vectors of length 128. The matrices $\mathbf{M_{r}}$ and $\mathbf{N_{r}}$ of the action inference model are of size $128 \times 128$. The action inference model has rank 2. We use human rule to collect demonstration covering the whole 500 states. The performance of the human rule is optimal.       

To analyze our hybrid RL with expert state sequences, we compare with the following methods: 
\begin{itemize}
	\item {\bf A2C: }  This method trains as a standard RL task, ignoring the expert state sequences.  Its configuration is the same as our A2C component.
	\item {\bf Behavioral cloning with duality action inference (BC-Dual): } This agent does not consider RL signals and only optimizes its policies by cloning the inferred actions on expert states.  The action inference model is the same as ours. 
	\item {\bf Imitation Learning (IL):  } Only this agent has access to the expert actions. This agent utilizes the expert actions to conduct behavioral cloning directly. No reinforcement learning signal is leveraged. 
\end{itemize}

To evaluate our action inference model, we additionally compare two variants where our action inference model is replaced by a multi-layer perceptron (MLP) as illustrated in Figure~\ref{fig:net1}. Similar to our action inference model, the states are first projected as vectors of length 128, then two state embeddings are concatenated and passed through two fully connected layers, and each layer has 128 units followed by ReLU nonlinearity. The total number of parameters of the MLP is close to ours.\footnote{The number of parameters of our action inference model is $128\times128\times2\times2$ and the MLP is roughly $256\times128+128\times128\times2$.} The variants replace our action inference model with MLP are:

\begin{itemize}
	\item {\bf A2C with MLP-based action inference(Hybrid-MLP):} It is the same as our proposed hybrid RL method except the action inference model is replaced by the MLP baseline.
    \item {\bf Behavioral cloning with MLP action inference (BC-MLP):} It is the same as {\bf Behavioral cloning with duality action inference} except the action inference model is replaced by the MLP baseline.
\end{itemize}

Each agent is trained and evaluated on 16 independent runs with different random seeds.

\begin{figure}[ht]
     \includegraphics[width=0.45\textwidth]{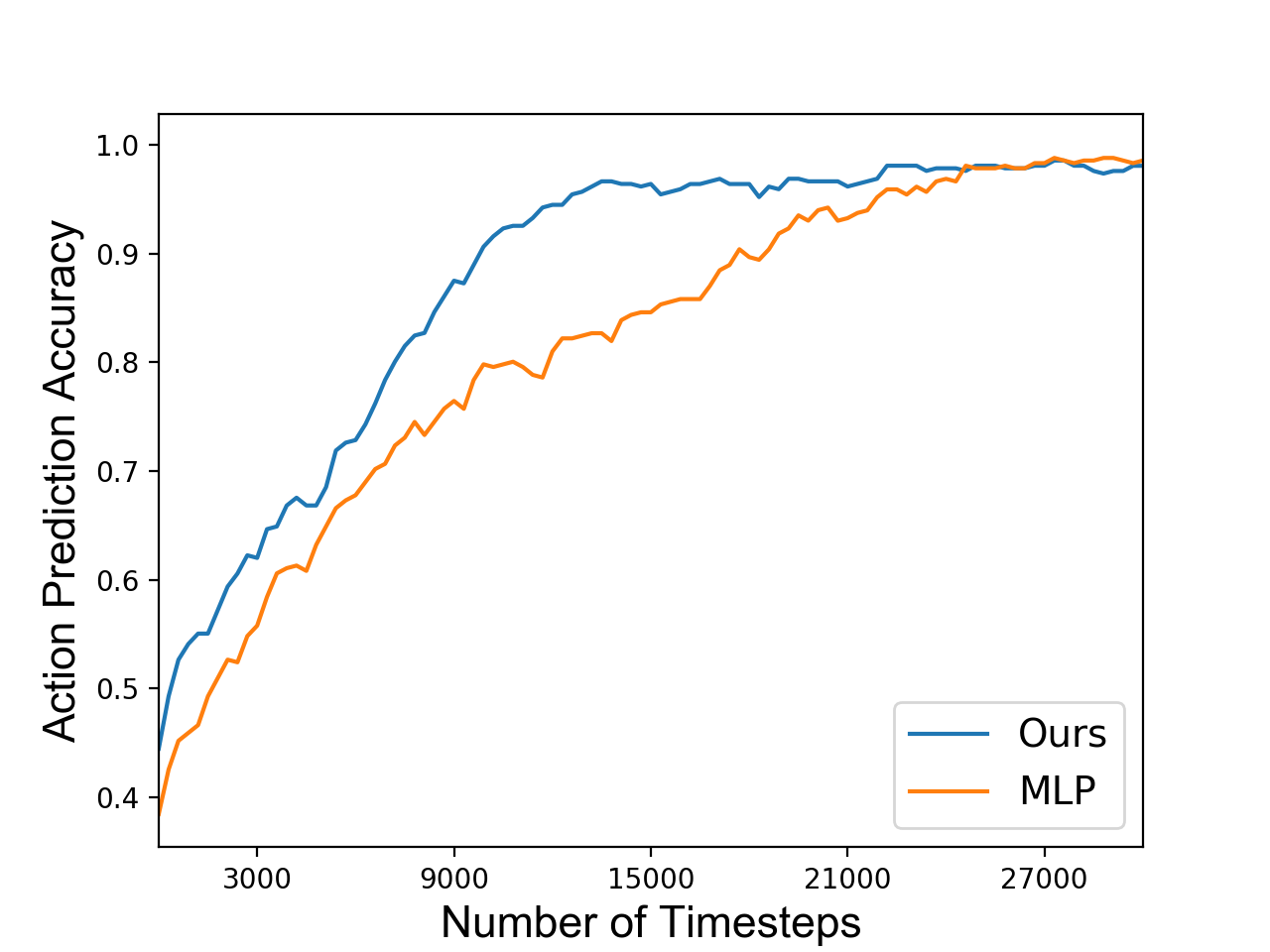}
\caption{Action prediction accuracy on the expert actions of our action inference model and the MLP-based action inference model in hybrid policy learning at different training time steps. \label{fig:taxi-pred}}
\end{figure}

\begin{figure*}[!t]
\centering
\begin{subfigure}[b]{0.245\textwidth}~
        \centering
        \includegraphics[width=0.95\textwidth]{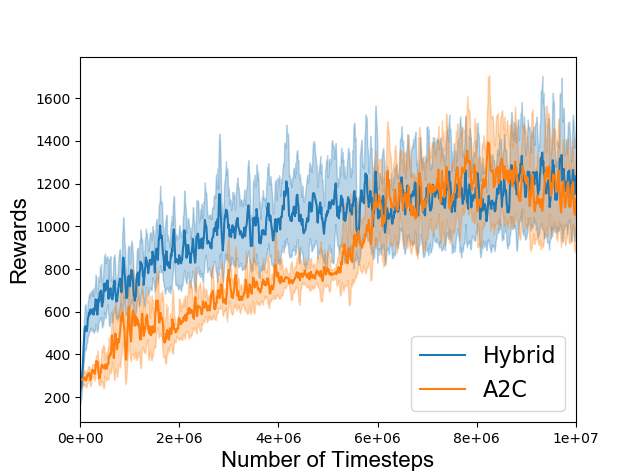}
\caption{Alien}
\end{subfigure}
\begin{subfigure}[b]{0.245\textwidth}~
        \centering
        \includegraphics[width=0.95\textwidth]{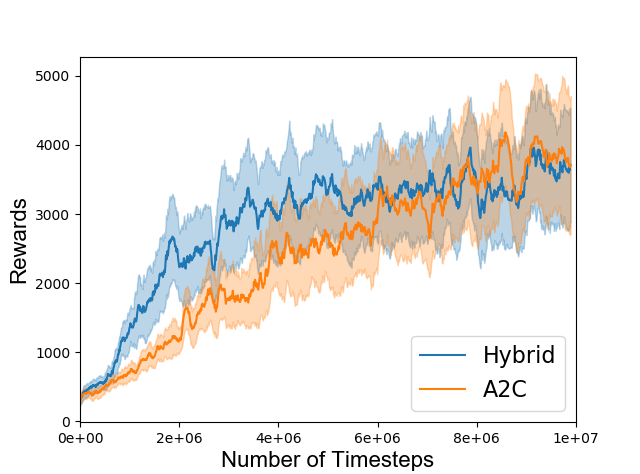}
\caption{BeamRider}
\end{subfigure}
\begin{subfigure}[b]{0.245\textwidth}~
        \centering
        \includegraphics[width=0.95\textwidth]{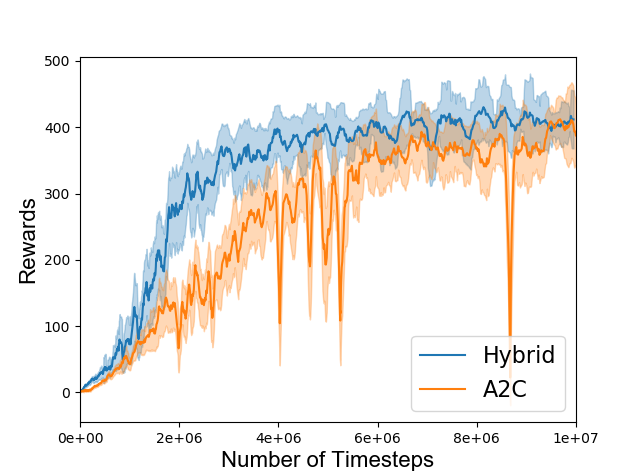}
\caption{Breakout}
\end{subfigure}
\begin{subfigure}[b]{0.245\textwidth}
        \centering
        \includegraphics[width=0.95\textwidth]{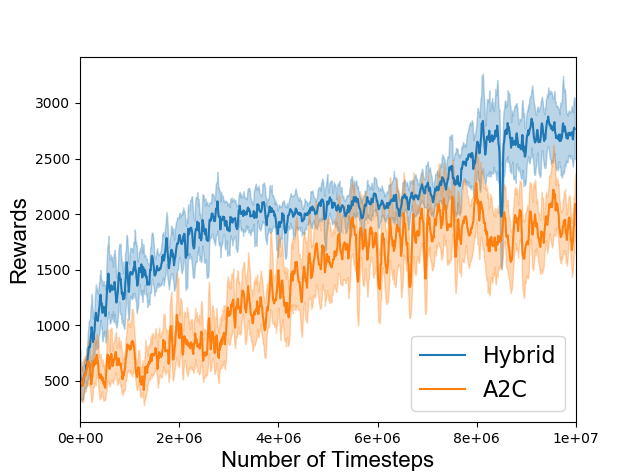}
\caption{MsPacman}
\end{subfigure}

\begin{subfigure}[b]{0.245\textwidth}~
        \centering
        \includegraphics[width=0.95\textwidth]{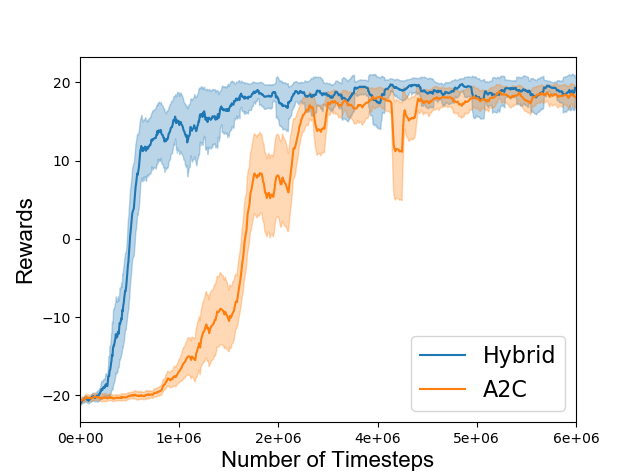}
\caption{Pong}
\end{subfigure}
\begin{subfigure}[b]{0.245\textwidth}~
        \centering
        \includegraphics[width=0.95\textwidth]{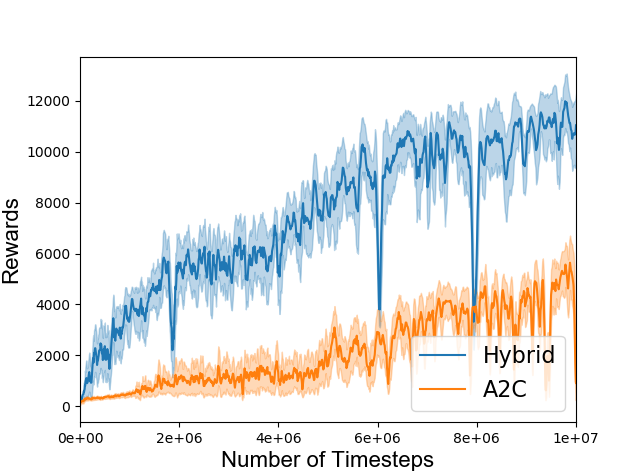}
\caption{Qbert}
\end{subfigure}
\begin{subfigure}[b]{0.245\textwidth}~
        \centering
        \includegraphics[width=0.95\textwidth]{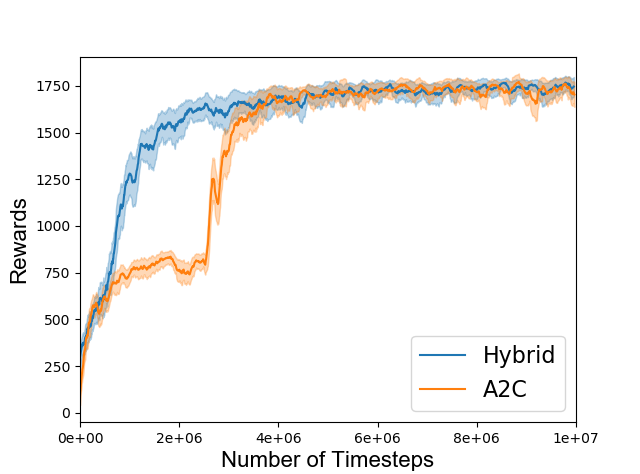}
\caption{Seaquest}
\end{subfigure}
\begin{subfigure}[b]{0.245\textwidth}
        \centering
        \includegraphics[width=0.95\textwidth]{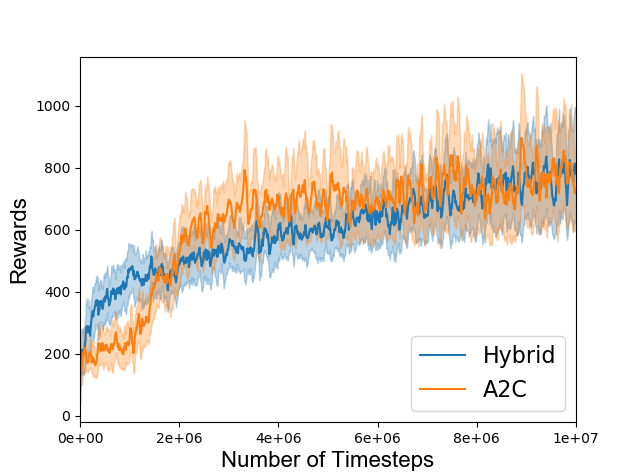}
\caption{SpaceInvaders}
\end{subfigure}

\caption{Learning curves of our method and the A2C baseline on eight Atari games. The figures show the average scores of the trained models with different number of image frames. The shadow regions represent one standard deviation. \label{fig:atari}}
\end{figure*}

\subsubsection{Experiment Results}  
Figure~\ref{fig:taxi-1} shows learning curves of different agents, and we make the following observations.  1) By comparing the hybrid agents with their pure imitation learning counterparts (Ours vs. BC-DUAL, Hybrid-MLP vs. BC-MLP), we see the hybrid agents have better performance. 
{Pure behavioral cloning  training signals only discriminate optimal vs. non-optimal actions, while the reward signals could help the agent to discriminate all actions, which could help to identify good actions to explore the environment. } With the help of RL signals, the distribution of training data for the action inference model changes such that it will learn faster on key states.
The results also show that, while the pure behavioral cloning agent improves rapidly at the beginning of learning, it takes much longer than the hybrid counterparts to reach the optimal policy.     2) By comparing our action inference model to its MLP counterparts (Ours vs. Hybrid-MLP, and BC-Dual vs. BC-MLP), the dual action inference model demonstrates performance advantages.  Since the dual model does not have any nonlinear transformations as MLP, our action inference model is  more data efficient in learning the environment dynamics and it is more robust to state distribution changes in learning as shown in Figure~\ref{fig:taxi-pred} showing the action prediction accuracy in Ours vs. Hybrid-MLP. 
3)  Both the hybrid agents outperform the pure A2C agents, which shows that the hybrid objective is effective in leveraging the expert state sequences to facilitate learning.

\paragraph{Effect of Ranks on Performance \& Parameter Reduction} Our low-rank tensor model efficiently reduces the parameter space while keeping good enough performance. First, the full parameter tensor without any low-rank approximation technique contains $500 \times 6 \times 500$ parameters for the Taxi domain ($|\mathcal{S}|=500$ and $|\mathcal{A}|=6$).
In comparison, our best-performing rank-2 model has a total number of parameters of $\vert \{\mathbf{M_{r}},\mathbf{N_{r}}\}_{r=\{1,2\}} \vert=(128 \times 128 \times 2) \times 2$, compressing the original tensor at a ratio of $4.37\%$.
Figure~\ref{fig:taxi-2} compared the learning curves with different ranks. Even the rank is set to 1, the performance is degenerated but the advantage over pure A2C agents still preserves. 
Setting a higher rank (R=4) does not improve the results and the learning curve is almost the same as the rank-2 model.

\paragraph{Robustness against Noise in Demonstrations} The above results demonstrate the performance advantage in an ideal setting where the expert state sequences cover the whole state space and the expert behaves optimally. We analyze the robustness of our agent against potential noise from the expert state sequences. We study two potential types of noise in expert state sequences: (1) Missing state ratio ($\eta$), namely the percentage of states that do not exist in the expert state sequences, and (2) Non-optimal action ratio ($\epsilon$), the percentage of state transitions caused by non-optimal actions. The performance of our agents for various values of $\eta$ and $\epsilon$ is summarized in Figure~\ref{fig:taxi-3}. By comparing the hybrid agents with the pure behavioral cloning agents, the hybrid agents are able to recover the optimal policies while the pure behavioral cloning agents get stuck at certain sub-optimal policies. This illustrates that the hybrid approach relies less on the optimality of the demonstrations. The figure also shows that non-optimal state transitions have a significant impact on the performance because the agent could get stuck in the Taxi domain, which could result in a sum of negative rewards until the maximum number of steps is reached. 

\subsection{Atari Games}

We evaluate our method on eight Atari games with machine generated state sequences to evaluate the scalability of our method. 

\subsubsection{Experiment Setup}

\paragraph{Model Architecture} We adopt the commonly used CNN architecture as in~\citet{mnih2015humanlevel} for Atari games. As shown in Figure~\ref{fig:net2}, the last four images are stacked in channel and rescaled to $84 \times 84$ as state input. The state encoding function is a four-layer convolutional neural network. The first hidden layer convolves 32 8$\times$8 filters with stride 4. The second layer convolves 64 4$\times$4 filters with stride 2. The third layer convolves 32 3$\times$3 filters with stride 1. The last layer of the state encoding function is fully-connected and consists of 512 output units . Each layer is followed by ReLU as nonlinearity. The last output vector is passed through a linear layer to generate the value estimates for the critic of A2C and through another linear layer followed by a softmax as the policy for the actor of A2C. Our action inference model shares the same state encoding CNN. The last output vector of length 512 is first passed through a linear layer with 128 output units. The matrices $\{\mathbf{M_{r}}, \mathbf{N_{r}}\}_{r}$ are all $128 \times 128$. The rank is set to be 8. 

We use pre-trained A2C agents with 5 million frames to generate 100 trajectories as demonstration state sequences. Each trajectory is terminated when one player life is lost. Similar to the Taxi domain, the agents are trained and evaluated on 16 simultaneous environments with different random seeds. 

\paragraph{Expert State Sampling Curriculum} Unlike the Taxi domain, the state mismatch between the demonstration and the agent at the beginning of the learning is significant. As the action inference model is only trained on the players' own state distribution, the demonstration states that are far from the agent's own experience could be wrongly labeled. Because of this, the pure behavioral cloning from demonstration agents (BC-) fail in achieving reasonable performance. On the other hand, learning from states that are far away from the agent's own state distribution is not immediately helpful. To mitigate such mismatch, we apply a curriculum in sampling the demonstration when optimizing the policies.  Specifically, we only sample from the first $K=10$ time steps of each trajectory at the beginning of learning. We gradually increase $K$ by 1 for approximate 8,000 frames.  In this way, the sampled state distribution of the demonstration matches better to the players' own experience. Furthermore, we only use the inferred actions after 100,000 frames to optimize the agent's policies when the action prediction model becomes reliable. 

\subsubsection{Experiment Results} The learning curves of our agent and the A2C baseline are shown in Figure~\ref{fig:atari}. Of all the eight games we evaluate, the hybrid agent is able to leverage expert demonstration to speed up learning on six games (Alien, BeamRider, Breakout, MsPacman, Pong and Qbert, Figure~\ref{fig:atari}(a-f)). For the other two games, A2C seems to stuck at Seaquest at a score of 1800, either the hybrid agent or the A2C agent is able to escape from that local minima; the action inference has the worst accuracy on the game SpaceInvaders. 

\section{Conclusion }
We have proposed an iterative learning paradigm to facilitate the problem of decision-making by utilizing demonstrations from  experts.   Differ from many previous approaches, we consider a  realistic and difficult setting that actions performed by the experts are unavailable.  To better make use of the state-only demonstrations, we propose to learn a novel tensor-based action inference model based on the agent's own experience.   The learned dynamics is further used to infer the missing actions from the expert demonstrations.  At last, a hybrid objective is proposed that improves the policy via imitation learning and A2C jointly.  The experiment results on eight Atari games and an illustrative Taxi domain  demonstrates the advantageous performances of our model against a set of baselines.  We also show that our model is robust against noisy expert state trajectories.

\bibliographystyle{aaai}
\bibliography{aaai2019.bib}

\end{document}